\documentclass[runningheads]{llncs}
\usepackage{graphicx}
\usepackage{algorithm} 
\usepackage{algpseudocode}
\usepackage{cite}
\usepackage{amsmath}
\usepackage{subfig}

\begin{document}
\title{Prototype of a robotic system to assist the learning process of English language with text-generation through DNN}
\titlerunning{Robotic system to assist the learning of English through DNN}
%
%
%

\author{Carlos Morales-Torres\inst{1}, Mario Campos-Soberanis\inst{1,2}, Diego Campos-Sobrino\inst{1,2}}
\institute{Universidad Politécnica de Yucatán, Ucú, Yucatán, México\\
\email{danniel.torres99@gmail.com} \\
\and
SoldAI Research, Mérida, Yucatán, México \\
\email{\{mcampos,dcampos\}@soldai.com}}

%
\maketitle              
\begin{abstract}
In the last ongoing years, there has been a significant ascending on the field of Natural Language Processing (NLP) for performing multiple tasks including English Language Teaching (ELT). An effective strategy to favor the learning process uses interactive devices to engage learners in their self-learning process. In this work, we present a working prototype of a humanoid robotic system to assist English language self-learners through text generation using Long Short Term Memory (LSTM) Neural Networks. The learners interact with the system using a Graphic User Interface that generates text according to the English level of the user. The experimentation was conducted using English learners and the results were measured accordingly to International English Language Testing System (IELTS) rubric. Preliminary results show an increment in the Grammatical Range of learners who interacted with the system.

\end{abstract}

\section{Introduction}

As Artificial Intelligence (AI) becomes more equipped to comprehend human communication, more institutions will adopt this technology for areas where Natural Language Processing (NLP) would make a difference. AI technology is already being used in smart home and office assistants, customer service, healthcare, and human robotics, among others.

There are multiple aspects of AI and NLP that generate the opportunity of having machines offering engaging, interactive capabilities. However, the current state of the art in NLP lacks reasoning and empathy capabilities, making complex interactions difficult. One way to exploit NLP technology engagement potential is the application of assistive technology. A particularly interesting field is the use of such systems in interactive robotics.

Humanoid robots are useful with tedious and risky errands for people, including tasks that can result in exhausting for human beings. Jobs that require a lot of concentration and feedback, like tutoring and guidance, can benefit from incorporating autonomous robotic systems to let the students interact with learning about a specific field. Robotic systems will require the capacity to understand human lexis to achieve these goals, making characteristic language handling more significant.

In the educational context, there are systems capable of teaching or assisting individuals in a self-learning process, such as Conversational Intelligent Tutoring Systems. However, they are still not optimal enough to automatically provide knowledge to help students in the learning process of a language without the need of human assistance \cite{cits2019}. Also, there have been interesting studies that show that interactive robotic systems are beneficial for learning \cite{systematic_review_eudcational_robotics_2019}. The previous characteristics devise a synergy opportunity of a robotic system that incorporates an NLP component to be helpful in the self-learning process \cite{12}.

This article presents a functional prototype of a robotic system to assist the English language learning process through text-generation using Deep Neural Networks (DNN). A humanoid robot was designed and manufactured to promote learners' engagement with the assisting tool. The interaction was conducted using a Graphical User Interface (GUI) incorporated in the robot. A text-generation component was included to allow the users to interact with the system and generate language using different English levels. The experimentation was conducted with English learners and measured using the International English Language Testing System (IELTS) rubric. Preliminary results show an improvement of the subjects' current English level through regular usage of the system. However, there is a need for further and deeper experimentation to generalize the findings in this work.
 
The article is structured as follows: Section 2 describes the state of the art of robotic systems implemented to assist self-learning; Section 3 presents the research methodology; Section 4 describes the experimental work carried out, presenting its results in Section 5. Finally, conclusions and lines of experimentation for future work are provided in Section 6.

\section{Background}

A humanoid robot is a robotic system capable of presenting similar features to resemble human anatomy. These robots are usually presented and utilized as a research tool in scientific fields aimed to understand the human body structure and behavior to build. It has been proposed that robotics will be helpful in various education scenarios\cite{systematic_review_eudcational_robotics_2019}.

Previous studies indicate that robotics is providing benefits as a teaching tool in particular in the STEM fields\cite{stem2005}, and English learning\cite{r-learning2009}. Robotic systems also provide a learning environment that seeks to improve the interdisciplinary process of learning, promoting the engagement of students in their learning activities \cite{robotics_multidisciplinary_2013, robotics_colaboration_2017}. There are examples where the use of a robot for assisting the learning process is appropriate to use in language skill development as it allows a richer interaction than digital platforms \cite{robots_in_education_2016, robotics_colaboration_2017}.  

A significant challenge to incorporate robots as a tool to assist the self-learning process of a language is to design an engaging experience tightly related to the language the learner is using. NLP is particularly well suited to close this gap. NLP has evolved from simple classification methods like logistic regression to more complex language statistical methods and DNN \cite{1}. Neural Networks are the dominant paradigm in NLP and have increased the research of end-to-end systems for understating human language, leading to complex applications as conversational chatbots \cite{18}. 

The current and approachable theory of already-existing NLP models makes extensive use of transformers, which are topologies that use an encoder-decoder architecture incorporating an attention mechanism\cite{attention_2020}. Many state of the art results make use of this architecture training with vast amounts of information. Models like BERT\cite{bert}, T5\cite{T5_2019} and GPT-3\cite{gpt3_what_is_good_for_2021} are examples of big transformers delivering state-of-the-art results for various NLP tasks. Nevertheless, the field of NLP is still underdeveloped in terms of using low data quantities to perform fine-tuning in big transformers models.

One way to deal with low quantity data for NLP tasks is using RNNs. These models are effective for predicting sequence analysis tasks \cite{Huang2015}, as they store the information for the current feature based on previous information, including within the model forecasting and conditioned output capabilities \cite{LSTM_forecast_2019}. 

Recurrent architectures learn the relative importance of different parts of the sequence; nevertheless, transformers substitute recurrent mechanisms with attention mechanisms \cite{attention_2020}, which allows the capture of longer size dependencies while reinforcing training.

There exist studies that favor traditional models like Conditional Random Fields (CRF) and LSTM networks over big transformers models in settings where the amount of data is not enough to perform fine-tuning, or the language specificity makes generalization difficult\cite{27, 28}. Additionally, LSTM runs faster, making it well suited for real-time systems interaction \cite{LSTM_Benchmark_2015}.

Language models (LM) also have been used for text-generation either using large transformers \cite{transformers_text_generation} or LSTM like in \cite{LSTM_language_models_2018, 23}. In this research, an LM is generated using an LSTM trained on a specific dataset, and it is used to predict the succeeding word. The predicted output word is then appended with the existing input words and given as new input. This process is continuously repeated by shifting the window to generate text.

In the presented work, a humanoid robotic system was designed and manufactured to help engage in the self-learning process of English language students. A text-generation module to expose users to a variety of vocabulary and sentences was developed, thorough the experimentation, selection, and fine-tuning of LSTM models, transformers, and encoder-decoder architectures. The best model is selected to perform text-generation using a lower seed-text as shown in \cite{LSTM_text_generation_2020}.

\section{Methodology}

This section presents the tools, methodologies, and development approaches used for corpus creation, text-generation module training, humanoid robotic system design, and the system integration to allow students to interact with it.  

\subsection{Corpus creation}

The dataset consisted on different English sentences divided into three categories: basic, intermediate, and advanced. A human expert IELTS evaluator assisted in the creation of sentences with different levels of English proficiency, considering variation in grammatical range and lexical resources according to each level.

The corpus is structured in sentences, divided by punctuation signs that are further cleaned and omitted to individual process words in the text-generation model. It contains 4,785 sentences and 150,000 words.

\subsection{Text generation module}

Most advanced models for text-generation make use of deep learning models, including LSTM networks and transformer architectures \cite{26}. Different DNN models were trained using the dataset described in the previous section to develop the text-generation component. The researched models were: Simple LSTM model, BERT fine-tuned model, Encoder-Decoder LSTM model, Bidirectional LSTM model.

\begin{figure}[h!]
  \centering
  \includegraphics[scale=0.9]{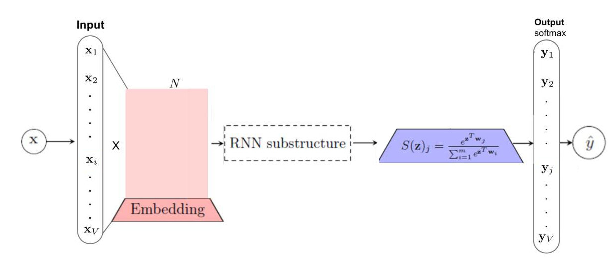}
  \caption{General architecture of the text-generation network}
  \label{fig:general_topology}
\end{figure}

To process the text, the input sentences were tokenized and passed through the input layer of each model, then to an embedding layer, and subsequently fed to the RNN substructure that processes the tokens. Finally a softmax layer is used to predict the probability of the next word. The general architecture of the networks are depicted in the figure \ref{fig:general_topology}

Each model was implemented using the Keras framework and trained using the same dataset split with 80\% for training and 20\% for testing. Also, at training time, a development set proportion of 10\% was used for Keras to compute validation loss and accuracy.

After experimenting with the mentioned models, the model with the best performance accuracy is selected and fine-tuned to perform the text-generation.

\subsection{Robotic system design}

The methodology used to design the humanoid robotic system consisted of three main phases: requirement definition, specification, and design.
In the requirement definition phase, an analysis of the functionality requirements of the robot was made, and the functional structures were defined. Then, through the specification stage, the robot and general guidelines for the project were carried on. In the design stage, specifications and guidelines were measured quantitatively, including the kinematics analysis and the definition of mechanical structures.

To favor student engagement with the robot, it was decided to use an anthropomorphic system bearing kinematics considerations. Regardless, the presented robotic system does not attempt to include mechanical components; the mechanical design was made to adopt mechanical actuators further to let the system move and increase interaction with users. The parameters that represent kinematics configuration in general terms were based on Denavitt Hardenberg\cite{denavithardenberg} motion equations. 


After the design stage was done, the system was drawn using the 3D drawing software fusion360. The manufacturing stage consists on printing and assembling a 3D sketch of the entire robotic system with the appropriate parameters obtained from the previous analysis.

\subsection{System implementation}
\begin{figure}[h!]
  \centering
  \includegraphics[scale=0.22]{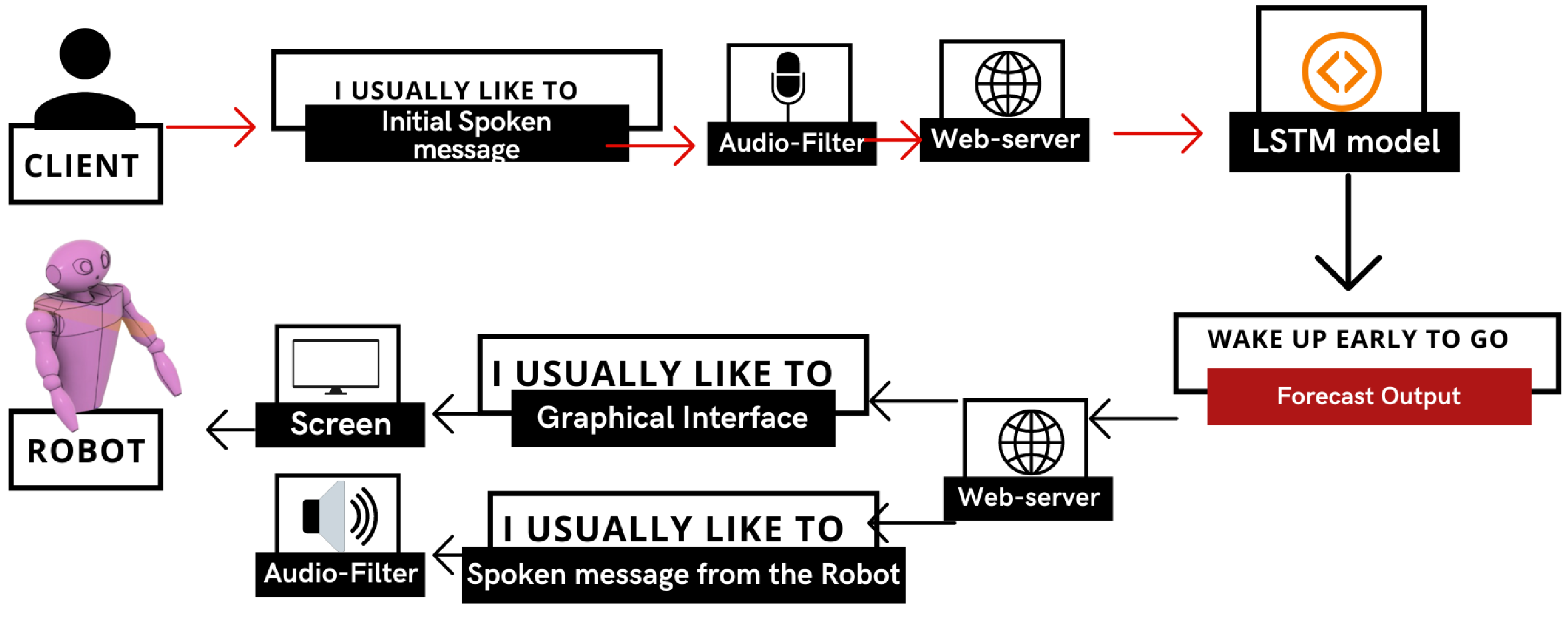}
  \caption{ System communication architecture. }
  \label{fig:communication_architecture.}
\end{figure}

The implementation includes an embedded system that captures the user's speech and uses Google's Text to Speech (TTS) web service to get the transcription of the user utterance. The embedded system sends the transcription to a web service implemented in Flask to consume the best text-generation model found in the experimentation. The implemented service uses the TTS transcription as a seed to predict the following text using a fixed number of 5 words. After the model predicts the text, the Flask server sends the predicted text to the embedded system using a webhook. The embedded system uses Google's Speech to Text (STT) service to generate an audio file with the predicted text and play it using a speaker. The system is attached to the robot's body, and the user initiates the interaction. Alternatively, a Graphic User Interface (GUI) was implemented using the Gradio library\cite{abid2019gradio}, which can consume the service using a tablet incorporated into the robot. The GUI was intended to include users with speech or hearing disabilities. The communication architecture is depicted in the figure \ref{fig:communication_architecture.}.

\section{Experimentation}
\label{section:experimentation}
This section shows the methods used for the text-generation module training, the manufacturing process of the robotic system, and the experimental process to measure system's effectivity to assist self-learning process for English students. The implemented mechanisms are illustrated, as described in section 3 of the document.

\subsection{Corpus data}
The corpus consisted of sentences with 3 different English levels: elemental (IELTS accuracy level 1-2.5),  pre-intermediate (IELTS accuracy level 3-4.5), and upper intermediate (IETLS accuracy level 6+). Each set contained different sequence-to-sequence compound-complex sentences. This was recommended by the IELTS evaluator to optimize three specific levels of English to tackle fluency levels in different scenarios. The corpus included 171,461 tokens, 150,356 words, and 4,785 sentences.

\subsection{Text-generation module}

The different models were trained using the corpus described in section 4.1 divided into random partitions for training, validation, and test. Four different models were trained: Simple LSTM model, BERT fine-tuned model, Encoder-Decoder LSTM model, and a Bidirectional LSTM model.
Each model was trained for 20 epochs, and the validation metrics were reported using the validation set. Different models were iterated using dropout regularization (\emph{dropout}) with different probability parameters. Once the best model was obtained in the validation set, it was evaluated in the test data to report the metrics presented in section 5.1.
The models were implemented using Tensorflow 2.0 and Keras on a Debian GNU/Linux 10 (buster) x86\_64 operating system, supplied with an 11 GB Nvidia GTX 1080 TI GPU.

After the first experiments were conducted the best performance model found was the Bidirectional LSTM measured in terms of accuracy and validation. Once the best model was found further experimentation was done using a grid search strategy to find the best hyper-parameters of the model resulting in the following topology: 
LSTM layer (100 units), Dropout Layer (0.6 drop rate), LSTM layer (100 units), Dense layer (100 units, ReLU activation), Dense layer (125 units, softmax activation).

The best parameters found were the following: Embedding vocabulary-size: 70, dropout layer: 0.6, activation function: softmax, trainable parameters: 180,275, loss function: categorical cross entropy, batch size: 150.

\subsection{Robotic system manufacturing}

The whole manufacturing design was approached under engineering methods to allow time-optimization and cost reductions to be considered. The process involves the following stages: Material Printing (Through a 3-D printing machine, segments from the material were printed to further treating and assembly), Material purification (Through chemical components, the segments of materials are purified through a specific epoxy designed to purify the material extracting impurities while adding brightness, Assembly of materials. (Through engineering glue, segments are assembled properly).

\begin{figure}[h!]
  \centering
  \includegraphics[scale=0.6]{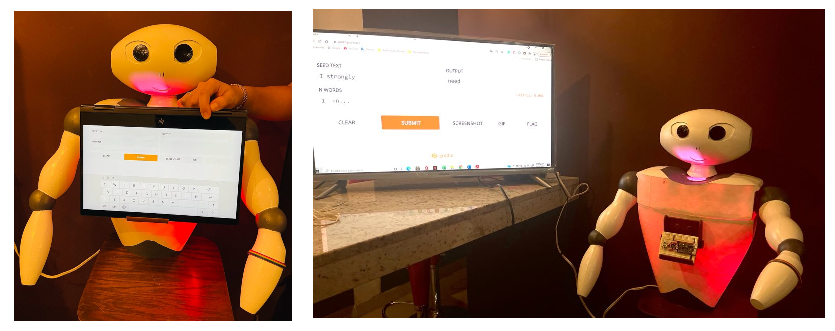}
  \caption{Robot configurations using embedded system and tablet. }
  \label{fig:robot_configurations}
\end{figure}

Each of the previous stages was divided in three segments: head-manufacturing segment, arm-manufacturing segment, body-manufacturing segment respecting each of the previously presented stages. Final configurations of the robot using the tablet and embedded system are presented in figure \ref{fig:robot_configurations}

\subsection{System Evaluation}

To evaluate the system's effectiveness to help learners, they were evaluated using an IELTS rubric before interacting with the system. After that, the learners were exposed to interact with the system for 5 days and a new evaluation using the same rubric was made to asses the performance of the students. The evaluation was conducted with three subjects, one for each English level in the corpus.

\section{Results}
\label{section:results}
This section shows the results obtained from the experimentation described in section \ref{section:experimentation}. The improvement of the subjects is analyzed from 250 recorded minutes of training with the system by each subject, including quantitative and qualitative evaluation from IELTS instructors. The system's performance was measured to determine the progress of the subjects.

\subsection{LSTM text-forecast model with encoded-decoded attention mechanism.}

Four different models were considered and evaluated to obtain the one with the best performance. Table \ref{table:models-accuracy} shows the accuracy obtained with the four different models when evaluated with the test dataset.

\begin{table}\label{table:models-accuracy}
\centering
\begin{tabular}{c c} 
    \hline
    Model Type & Accuracy  \\ [0.5ex] 
    \hline\hline
    Simple LSTM  & 80\% \\ 
    \hline
    BERT fine tuned & 80\% \\
    \hline
    Encoder-Decoder LSTM & 89\% \\
    \hline
    Bidirectional LSTM & 95\%\\
    \hline
\end{tabular}
 \caption{Model accuracy results.}
\label{table:1}
\end{table}

\begin{figure}[h!]
\subfloat{}{\includegraphics[width=0.45\textwidth]{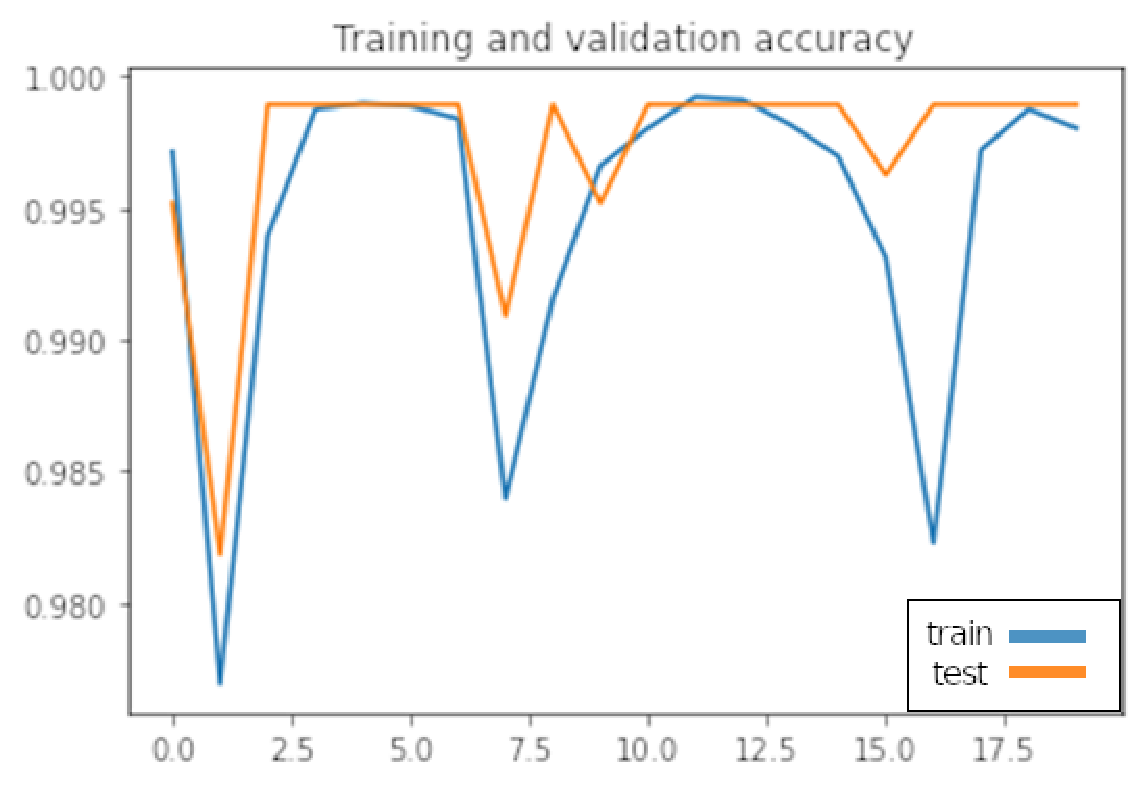}\label{fig:accuracy-loss-a}}
\hspace{0.08\textwidth}
\subfloat{}{\includegraphics[width=0.45\textwidth]{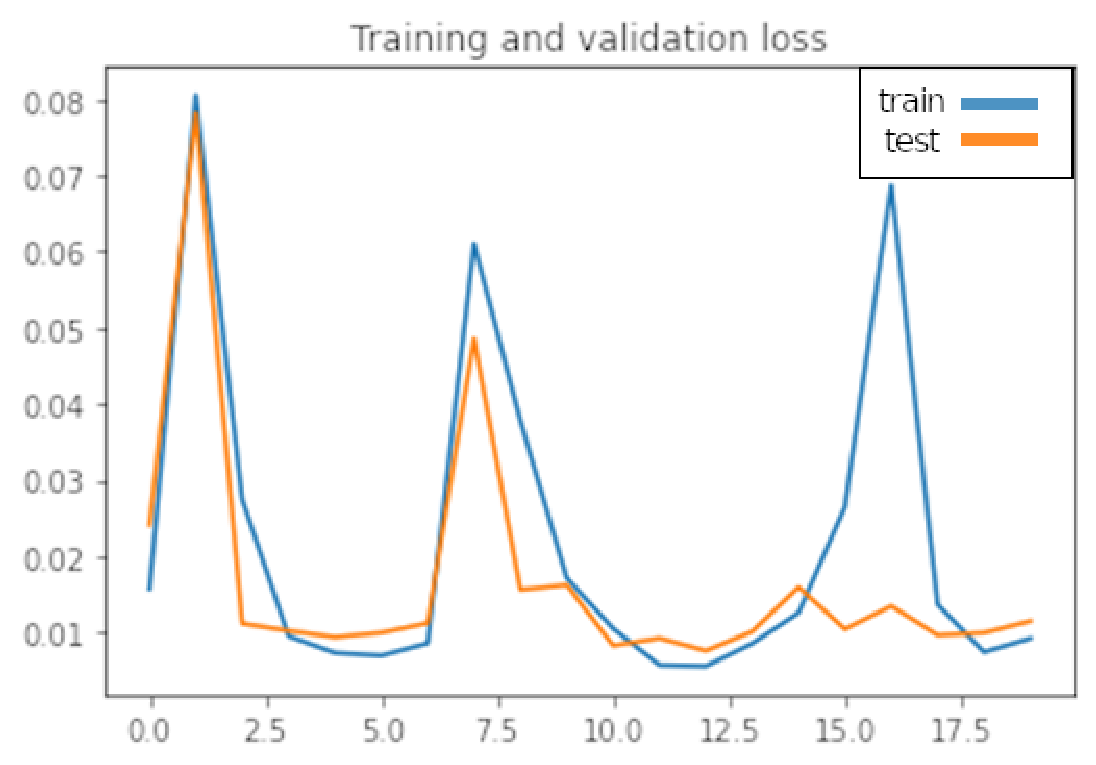}\label{fig:accuracy-loss-b}}
\caption{Accuracy and loss validation for Bidirectional LSTM model}
\label{fig:accuracy-loss}
\end{figure}

The most suitable model that provided results to be used on experimental subjects was the Bidirectional LSTM model. Figure \ref{fig:accuracy-loss} shows the training accuracy and loss for the 20 epochs of training of the Bidirectional LSTM model.

\subsection{Fluency improvement on subjects}

This section presents the outcome for the fluency analysis in each of the three experimental subjects after 250 minutes of interaction (50 minutes per day for five consecutive days) with the robotic system. 

The grammatical range and accuracy and marked by using a determined number of grammatical structures (6 types) in a percentage rate of accuracy and error-mistake (1-100\%). The assigned instructors included the number of grammatical sentence usage in terms of accuracy percentage. 

\begin{figure}[h!]
  \centering
  \includegraphics[width=1\textwidth]{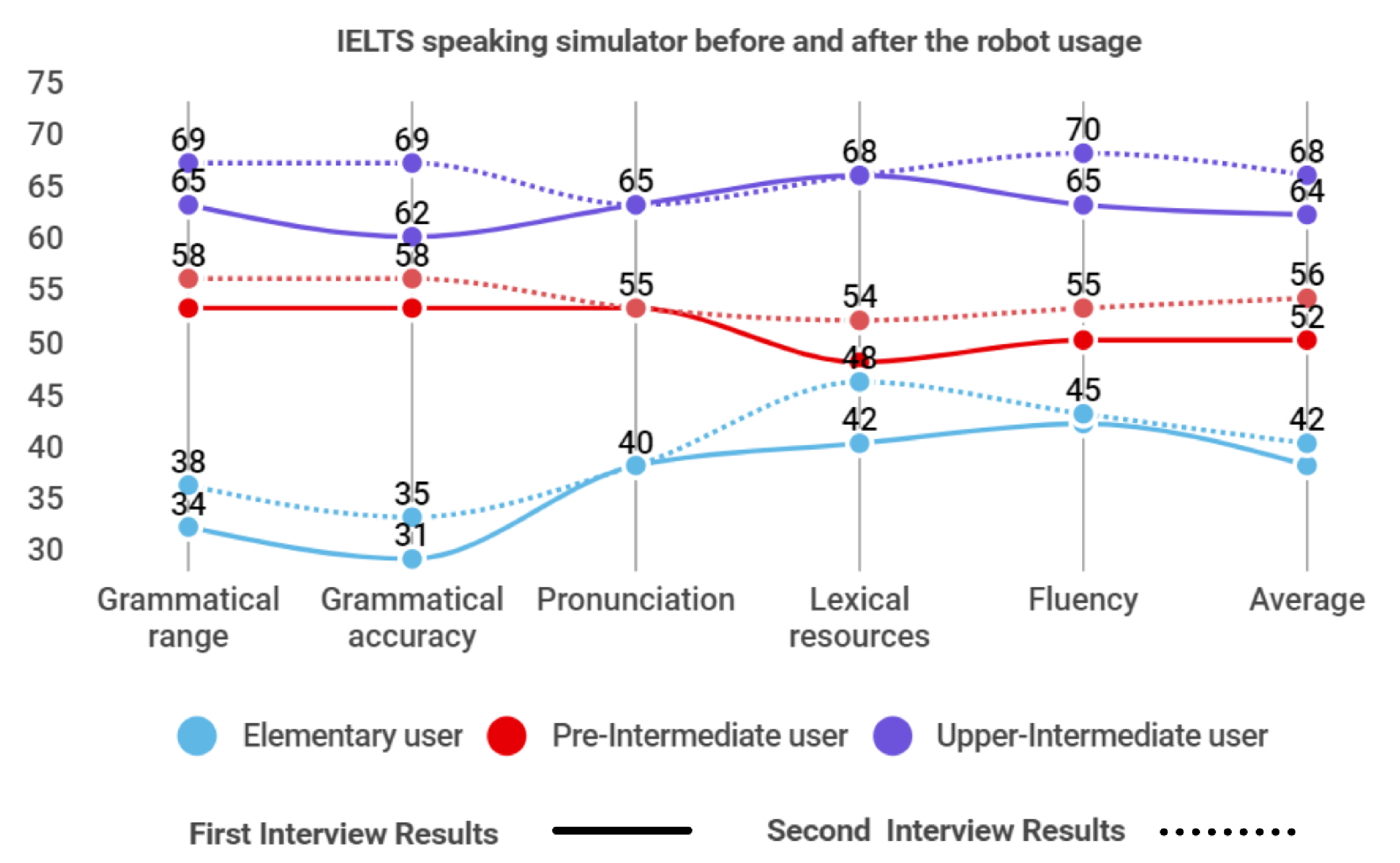}
  \caption{IELTS metrics comparison before and after training.}
  \label{fig:ielts-results}
\end{figure}

After elementary training, an increase in grammatical range and accuracy, lexical resources, and fluency is observed, while pronunciation and language-idiomatic terminology doesn't show improvement.  From the pre-intermediate level training, a sustained increase overall dimensions was observed, except for pronunciation. The upper-intermediate level attempted to evaluate fully understanding of complex ideas generated from the advanced corpus previously trained. The idea is to oversee a different set of more compound-complex sentences generated by the robotic system. The results before and after the training are showed in figure \ref{fig:ielts-results}.

\subsection{Qualitative results}

\begin{figure}[h!]
  \centering
  \includegraphics[width=1.0\textwidth]{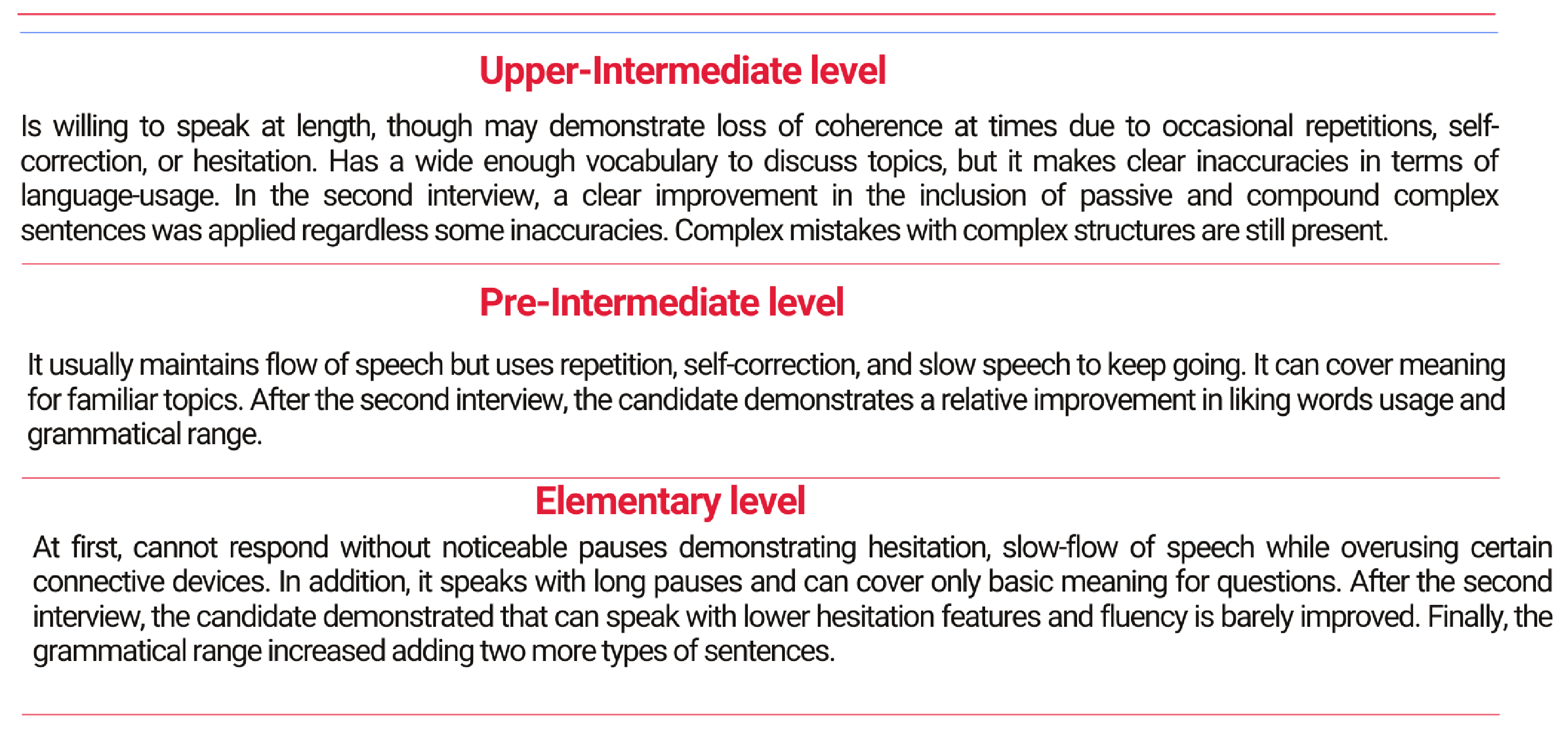}
  \caption{Qualitative feedback from IELTS instructor after training.}
  \label{fig:ielts-qualitative-feedback}
\end{figure}

The qualitative data obtained in this section was collected from IELTS instructors who evaluated and listened a set of questions from one specific context of coherence for each subject, to determine a mark in grammatical range and accuracy based on IELTS rubric. Finally, instructors who listened the same ideas in the second interview attached written feedback shown in the figure \ref{fig:ielts-qualitative-feedback}.

The results express that the instructors perceived noticeable enhancement in the English abilities of the subjects after the interaction with the robot.


\section{Conclusions and future work}

This work presented the design, development, and manufacturing of a humanoid robotic system to assist English language students in a self-learning process. The robotic system was developed using a three-phase methodology (requirement analysis, specification, and design) which yields good results since the system is articulated and ready to add further interaction using actuators. 

Various models were tested to implement the text-generation module; a particularly interesting observation is related to the relatively poor results (80\% accuracy) obtained when using a fine-tuned BERT model. This occurs due to the relatively small amount of data used to perform the fine-tuning; in this regard, the bidirectional LSTM model performs better, achieving a 95\% of accuracy in the test set.

The bidirectional LSTM text-generation model was useful to predict text using a seed given by the user; nevertheless, noticeable irregular fluctuations were reported on the validation accuracy and loss chart, which can be produced from irregularities in the English levels used within the corpus.

The experimentation was carried on with three English students of elementary, pre-intermediate, and upper-intermediate English levels, and their progress was measured according to the IELTS rubric. After 250 hours of training, comparative results demonstrated an average improvement of 4\% in their grammatical range, 4\% in grammatical accuracy, and 3.33 \% in their fluency. No difference was observed in their pronunciation abilities.

Quantitative and qualitative data obtained from the experimentation depicted a positive result on how a robotic system can provide aid while tackling a specific ability from a foreign language. In this case, the main improvements were reported in terms of fluency and grammatical range skills. Qualitative results show a favorable opinion both from IELTS instructors and students. In general, they perceived the system as a beneficial tool for the progress of the students.

The experimental results were limited by time constraints and the reduced number of subjects, so further research is needed to generalize the observed results.

The future work regarding this project includes: robust experimentation using more subjects and more structured training sessions, revision of other learning techniques and the overall effect on the English language improvement, experiment with variations on the composition of the corpus to measure its impact in the learning process. Also, interesting research can be conducted regarding pronunciation improvement using a more controlled spoken interaction with the users and the effect of dynamic movement adding actuators to the robot and measuring the impact in the self-learning process.



\bibliographystyle{splncs04}
\bibliography{references}
%
%
%
%
%
%
%
\end{document}